# Pouring Sequence Prediction using Recurrent Neural Network

Rahul Paul
Computer Science and Engineering Department
University of South Florida, Tampa, USA.

*Abstract*— **Human does their daily activity and cooking by teaching and imitating with the help of their vision and understanding of the difference between materials. Teaching a robot to do coking and daily work is difficult because of variation in environment, handling objects at different states etc. Pouring is a simple human daily life activity. In this paper, an approach to get pouring sequences were analyzed for determining the velocity of pouring and weight of the container. Then recurrent neural network (RNN) was used to build a neural network to learn that complex sequence and predict for unseen pouring sequences. Dynamic time warping (DTW) was used to evaluate the prediction performance of the trained model.**

*Keywords—Robot, pouring, Recurrent neural network, Dynamic time warping*

## I. INTRODUCTION

The automated industrial robots handle manufacturing, workload management, packaging etc. quite effectively with high precision than a human. In contrary, the everyday tasks, which are easier to handle by the human even though there is difference of environment, situation and material between the two events happened at two different time points. Teaching a robot to perform daily human life tasks is hard to fulfill. To execute tasks in daily environment, the robots need to be trained on different parameters e.g. weight of a material, state of a material etc. Researchers are working to generalize the robots to become useful in daily environments and the method of teaching a robot by demonstration and examples instead of machine commands is known as, Programming by Demonstration (PbD) [1]. In my previous work [2], I analyzed grasping of robots using objects at different states. In my current work, beyond rigid body manipulation, a simple daily task of pouring was analyzed based on motion trajectory and demonstration.

Though, pouring is a simple daily life task but it has many challenges to handle in contrast with solid objects. A human uses his vision to understand whether the container which was using to pour is empty or not or whether the container requires more rotation. The container as well as the liquid to pour can be different from time to time. For a robot to learn the pouring mechanism is difficult, e.g. the liquid to pour might be transparent or opaque, so by using the vision to analyze the liquid and pouring is difficult. Sometimes in some cooking or actions you may need different amount of water, so judge the amount of water based on the weight of the container, rotation required and density of the liquid is very difficult as well. Along with that, a container is used by the robot for pouring. The material, size and shape of the container may be different as well. For a pouring task, a force is applied to start and stop the mechanism. In this study, my objective is to analyze pouring motion trajectory analyze using the force feedback.

Imitation learning [3] is one of the earliest trajectory analysis approach for robots to imitate movement and motion of a human. But there are certain problems related with this kind of learning, e.g., quality of learning is heavily depended on the teacher and some work (grasping or pouring) can't be learn through imitation. Dynamical movement primitives (DMP) is another popular framework for obtaining motion trajectory [4]. This work was inspired by finding a complex motor actions that is adaptably accustomed without changing the parameters manually. Machine learning and statistical learning mechanisms were used to teach an autonomous dynamical system. Each DMP block is a non-linear dynamical system, whereas the previously proposed approaches weren't, that's the main distinction between DMPs and previous proposed approaches. DMPs can be subdivided into two categories: rhythmic (e.g. walking movement etc.) and discrete (e.g. tennis swing). A point attractor or a limit cycle is utilized as a base system for a DMP block of a discrete or rhythmic DMP respectively. DMP system had been utilized effectively for analyzing various trajectory movements, e.g., tennis rackets swing [5], walking [6], rhythmic drumming movement [7] etc.

Kulvicius [8] proposed a tightly coupled dual agent system based on DMP using a predictive learning approach along with sensory feedback. By adding the sensor, he showed that, both agents leant to co-operate. Based on DMP framework, Interaction primitives (IP), a human-robot interaction scenario was proposed by Amor [9]. In this study, demonstration from two interactive humans were used to train a IP. A learned IP was utilized by a robot to interact with a human. This study also considered timing, a variable parameter, between different persons and current mood or fatigue. Dynamic time warping (DTW) was utilized to analyze the performance of a learned IP in an unseen environment and situations. The disadvantage of the proposed IP method was the different interaction patterns between robot and human. To overcome this disadvantage, Ewerton [10] utilized the idea of IP, and proposed Mixture of Interaction Primitives based on Gaussian Mixture Model (GMM). As a result, different interaction patterns between human and robot could be modeled and a non-linear correlation could be established.

Another approach to model task trajectory and reproduction is based on GMM and Gaussian Mixture regression (GMR) respectively. Rubin [11] presented an algorithm to compute maximum likelihood function iteratively from an incomplete observations data using Expectation Minimization algorithm. A probabilistic PbD framework were presented by Calinon [12] for solving the task constraint and to utilize the learnt knowledge to various situations. In this paper, they combined Gaussian probabilistic representation with jacobian based solution to inverse kinematics. This approach allowed simultaneously handle constraints on multiple objects for different architectures. Calinon [13] proposed a robust model learning through human demonstrations. GMM and Hidden Markov Model (HMM) were utilized to point redundancies among multiple demonstrations and construct a time independent model using human demonstrations. Caldwell [14] showed that the gesture representation in robots could improve by using dynamical system and statistics using a GMR approach altogether. Billard [15] proposed an iterative algorithm called Binary Merging (BM) for a time independent autonomous dynamical system using distributions from mixture of Gaussian. The model leant through human demonstration for 2 degree of freedom robots and validated over 2D human motions library. Another paper by Billard [16,17] presented an approach to learn arbitrary non-linear autonomous motions from human demonstrations. They proposed Stable Estimator of Dynamical Systems (SEDS) that optimized parameters for GMM. For an operational space, this approach helped the robot to execute a task from any point, while keeping the motion as like the teaching demonstrations.

Principal component analysis (PCA) is another approach for motion generation and analyze for the robots. Mataric [18] analyzed the human motions and applied PCA to extract movement primitives as joint trajectories and applied them to generate and classify robotic motion. Park [19] proposed a framework for movement and learning of robot which integrated dynamic models, movement storage and optimization with a goal to achieve motion like a human. By using PCA they stored each base movement primitive as joint trajectory. Later these base primitives were combined linearly to achieve better human-like motion. Min [20] presented a new deformable motion model for human motion modeling and analyze. They applied statistical analysis using PCA to a demonstrated human motion data and constructed a deformable model of low dimension which helped in generating and identifying motions with great ease. Functional PCA (fPCA) is an improvement over PCA and uses statistical method to investigate the modes of variation in functional data with a goal to find low dimensional subspace of curves while capturing all the characteristic patterns [21]. Huang [22] presented a novel trajectory generating algorithm by extracting motion harmonics from demonstrated motion using functional analysis and then took constraint from user to generate new motions that corresponded to the demonstrated motions.

Recurrent neural network (RNN) is a state of art neural network model for analyzing and recognizing patterns from sequential data. One of the earliest study of RNN over the dynamic data was done by Han [23]. Reinhart [24] proposed a robot control framework trained on forward and reverse motion for movement generation using RNN. They also found out that, by changing the attractor states with the test target input, allowed the model to a wide range of unseen targets. Graves [25] used Long Short Term Memory (LSTM) a variant of RNN to generate both discrete and real complex sequence with long-range structure by predicting one single data point at any given time. Thus, even the cursive handwriting could be generated using their trained model. Kinematics, dynamics and trajectory analysis are needed for predicting robot's motion. Tang [26] utilized a RNN to solve non-linear functions and adapting to the unfamiliar trajectories using a virtual 6 degrees of freedom robot. Kim [27] investigated the robot's motion in tele-operation system. RNN was utilized for obtaining features from sequential data and by training the RNN with the features motion was generated which was comparable with human motion. Trischler [28] proposed a generalized dynamic system algorithm for the synthesis of RNN. By constructing a RNN using their approach would reproduce the original model's dynamics.

Liquid pouring is difficult task to learn for a robot even though it's a daily task for human. Pan [29] proposed optimization based receding horizon planner guided by machine learning model for the liquid pouring problem. Receding horizon planar gave both liquid overflow curve and prior mean trajectory. Yamaguchi [30] explored stereo vision to identify liquid flow as 3D point cloud. Optical flow algorithm was used to detect the flow and thus 3D information was generated by using stereo camera to learn the dynamical flow model. Brandl [31] tried to compute geometric parameters by warping of known objects for matching the shapes which would eventually helped the robot to identify different objects. This paper tries to follow the approach mention by Huang [39] and uses the same dataset for analysis.

The paper is as follows. Section II reviews the dataset. In section III the fundamentals of LSTM and RNN are reviewed. Section IV describes the approach taken and results over unknown test cases. Section V discuss the performance of the approach and concludes the study.

## II. DATASET

Six different cups (pour from container), ten different containers (pour to container), one ATI ini40 force and torque (FT) sensor and one Polhemus Patriot motion tracker was used for data collection procedure. All cups and containers were mutually distinct. The motion tracker recorded (x, y, z, yaw, pitch, roll) at 60 Hz whereas FT sensor tracked ($f_x$, $f_y$, $f_z$, $T_x$,

$T_y$, $T_z$) at 1 kHz. 3D adapters were used to connect the cup, FT and motion tracker as shown in Figure 1.

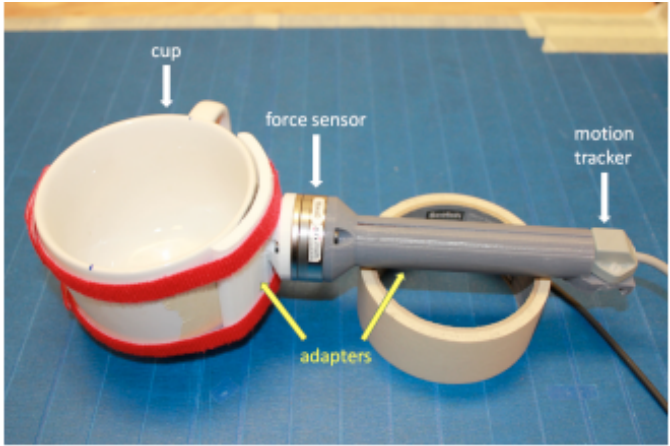

Fig. 1. Experimental setup for the pouring task

Three different materials were used to pour e.g. water, beans and ice. An empty reading was taken without putting anything in the cup to analyze the required force. 500 FT samples were taken in 0.5 seconds and then by averaging the initial force measurement was done. Similarly, for each pouring case, an initial trial reading just before the pouring and a final reading just after the trial reading with or without material was obtained. The sensed force was defined as,

$$f = \sqrt{f^2_x + f^2_y + f^2_z}$$

The training dataset consisted of 1307 motion sequences and their corresponding weight measurement. Each motion sequence had 8 associating attributes, so in total there were 10 attributes.

| | |
|---|---|
| $\theta_t$ | rotation angle at time $t$ (degree) |
| $f_t$ | weight at time $t$ (lbf) |
| $f_{init}$ | weight before pouring (lbf) |
| $f_{empty}$ | weight while cup is empty (lbf) |
| $f_{final}$ | weight after pouring (lbf) |
| $d_{cup}$ | diameter of the receiving cup (mm) |
| $h_{cup}$ | height of the receiving cup (mm) |
| $d_{ctn}$ | diameter of the pouring cup (mm) |
| $h_{ctn}$ | height of the pouring cup (mm) |
| $\rho$ | material density / water density (unitless) |

Only features $\theta_t$ and $f_t$ changed with time, other 8 features were kept constant throughout an entire sequence. An unseen dataset of 289 motion sequences were used to evaluate the performance of the trained model. Depends on the angle and pouring material, motion sequences can have different lengths. So, we padded the smaller sequences with zeros to make equal dimension to all the 10 features. $f_t$ was used as the target label and other nine features were used as input for the model.

## III. RECURRENT NEURAL NETWORK, LSTM AND GRU

Recurrent neural networks [32] are a type of neural network which is using to analyze time series prediction, sequence analysis etc. RNN considers learning methodology of a human brain: remembering or storing previously learned events and start learning new things or events from the previously learned events. RNNs are called as 'Recurrent' because they execute the same task for all the elements of a sequence and their output of the present time point t depends on the previous time point t-1. If we try to unroll the loop-like structure of a RNN, we will get a chain like normal neural network block structure (shown in Figure 2), where the information is processed and passed to a successor from a predecessor block.

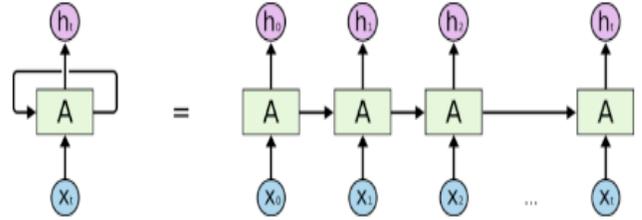

Fig. 2. Example of a RNN block

Recent years, the RNNs are using extensively in various fields e.g., time series prediction, speech recognition, text generation, language modeling, etc. Even though the RNNs are quite popular, they have a problem of vanishing gradient [33]. Backpropagation method is utilized to train a RNN through time, and when we unfold a RNN into multiple chain-like deep neural network, the gradient passes back through many time stamps. As a result, the gradient become very small and weight of the network doesn't change, which may stop the training of the network.

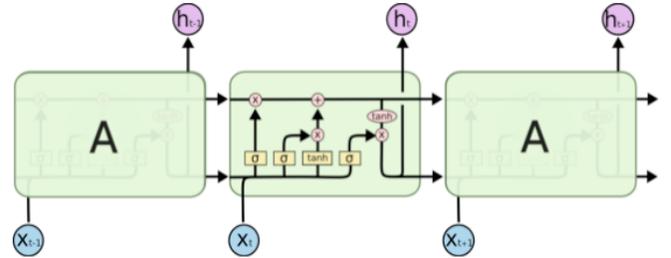

Fig. 3. Example of a LSTM block

Long Short Term Memory (LSTM) [34] is a variant of RNN that is used to avoid v vanishing gradient problem. LSTM also has a chain-like structure but each of the repeating block has different inside structure (shown in Figure 3). Each block consists of 4 neural network layers, in three of the layers there are sigmoid activation function and the other one has tanh activation function. The sigmoid layers are connected with three gates. The sigmoid layer output has a range in between 0 to 1. Zero means nothing will go through the gate, and 1 means all will go through the gate. These gates help LSTM to control flow of information and cell state. The first gate is the forget gate (Figure 4), which is responsible for removing and deleting information from a block. When an information is no longer needed for the LSTM for further processing of sequence, that information is removed by this forget gate. Second gate is called Input gate, which is accountable for adding new information in the block. Output gate is the last gate, whose work is to select useful information from the current block and send that to the next block for further processing.

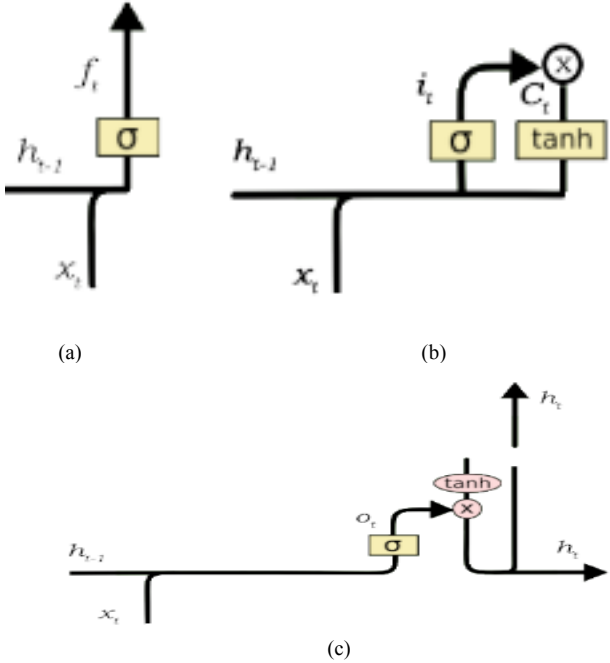

Fig. 4. LSTM's gates (a) Forget gate, (b) Input gate and (c) output gate

TABLE I. DIFFERENT MODELS AND PARAMTERS

| Model 1 | Model 2 |
|---|---|
| LSTM_1 =16 | GRU_1 =16 |
| LSTM_2=16 | GRU_2=16 |
| Dropout_1 =0.5 | Dropout_1 =0.5 |
| LSTM_3 =16 | GRU_3 =16 |
| LSTM_4 =16 | GRU_4 =16 |
| Dropout_2 =0.5 | Dropout_2 =0.5 |
| Dense _1 =1 | Dense_1=1 |
| Sigmoid/linear/tanh | Sigmoid/linear/tanh |

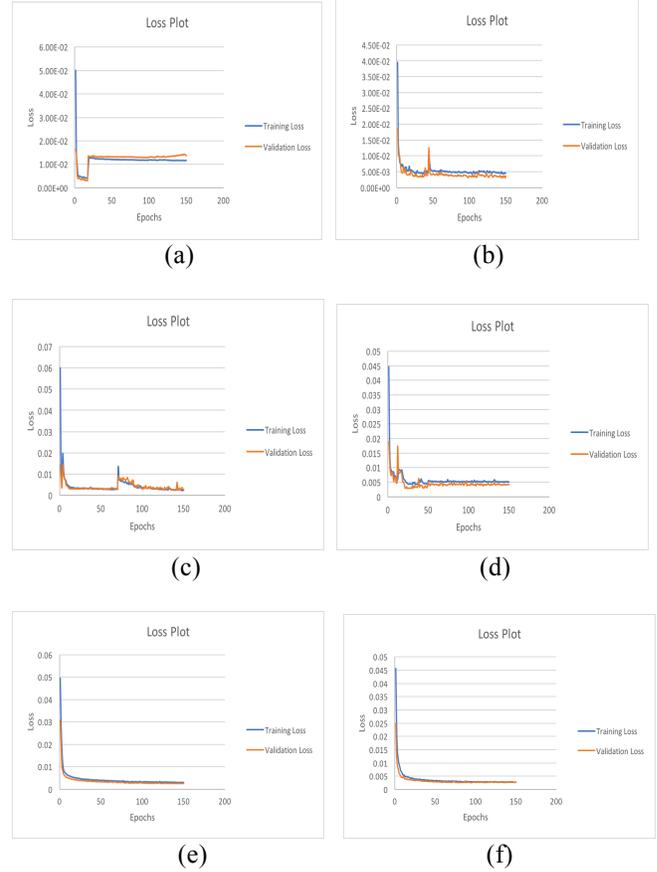

Fig. 5. Loss function plots (a) LSTM-Sigmoid, (b) LSTM-linear (c) GRU-sigmoid (d) GRU-linear ,(e) LSTM-tanh and (f) GRU-tanh

## IV. EXPERIMENTS AND RESULTS

In this section, we will analyze the experiment procedure and results obtained. We utilized LSTM and GRU to build our model. Though we have experimented with different setting of batch size and learning rate and activation functions, in this paper I am presenting six different models shown in Table 1. In the LSTM (GRU) blocks, tanh activation function was used, whereas in the final layer sigmoid, tanh or linear activation were used. Each of these models were trained using adam [35] gradient descent algorithm for 150 epochs with learning rate 0.01. Mean squared error was used as the loss function. Dropouts [36] of 0.5 were added after two LSTM/GRU layers to reduce overfitting of the training model. The training dataset was separated into 70% training and out the remaining 30% data-points, 90% data were used for validation and 10% to test the model. The final model was tested with an unseen test dataset. The loss plots of these models are shown in Figure 5.

We evaluated the prediction performance of our models on the unseen test data using Dynamic time warping (DTW) [37]. DTW is one of the algorithms to evaluate the resemblance of two sequences in time series analysis domain. DTW algorithm can be applied to two motion sequences with different speed. FastDTW [38] toolbox was used for calculating DTW. DTW diagram of the models are shown in Figure 6. From the DTW plots, we saw that, the actual and predicted values are not very similar but there was some similarity between them. For better analysis, prediction plots of six motion sequences namely, 286,18,10,171,267,203 from different models are shown in Figure 7.

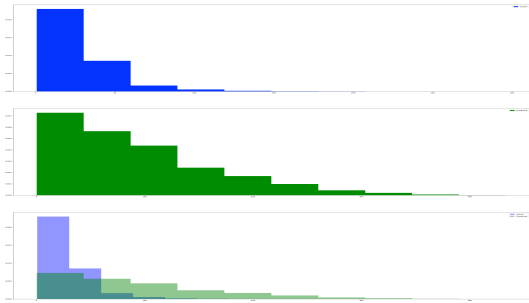

(a)

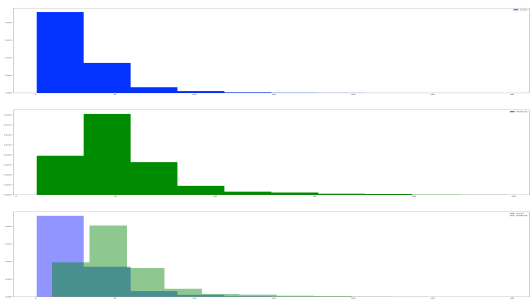

(b)

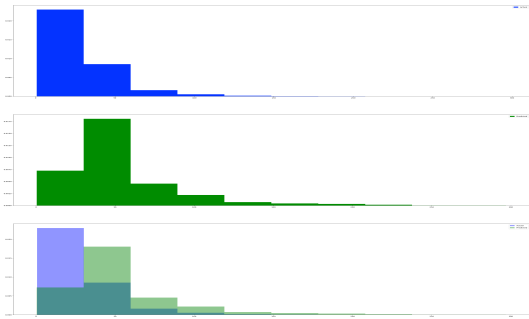

(c)

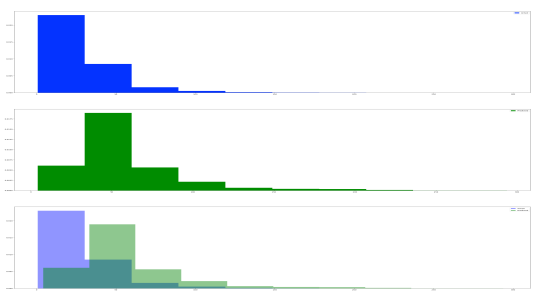

(d)

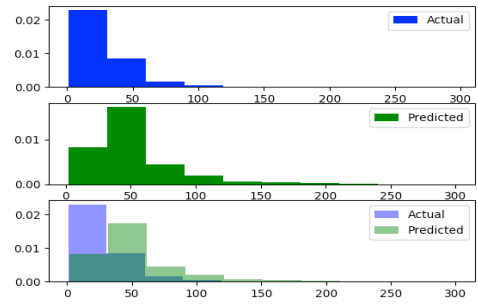

(e)

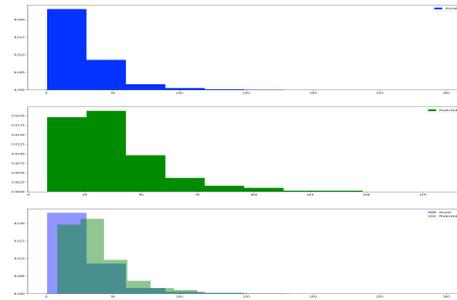

(f)

Fig. 6.  DTW plots (a) LSTM-Sigmoid, (b) LSTM-linear (c) GRU-sigmoid (d) GRU-linear ,(e) LSTM-tanh and (f) GRU-tanh

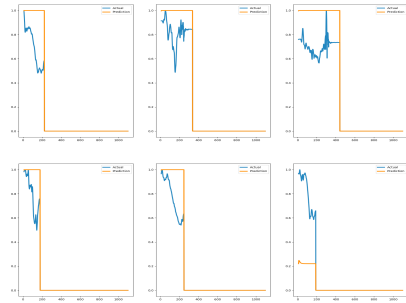

(a)

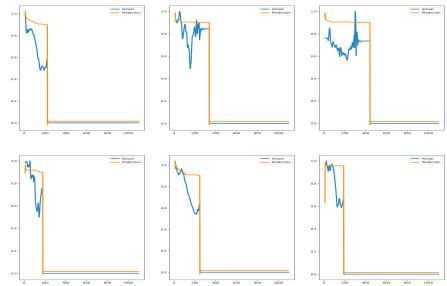

(b)

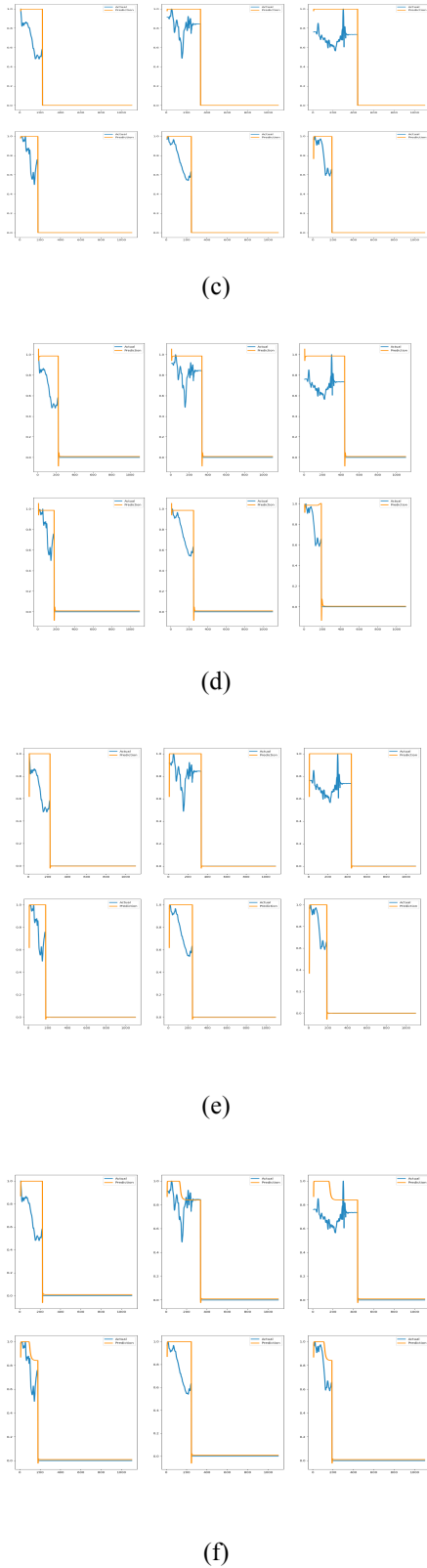

Fig. 7. Performance plots (a) LSTM-Sigmoid, (b) LSTM-linear (c) GRU-sigmoid (d) GRU-linear ,(e) LSTM-tanh and (f) GRU-tanh

## V. Discussions and conclusions

Pouring is simple daily human task. For this paper, trying we tried to learn the pouring approach by using recurrent neural network. Reduction of weight in the pouring container is used as label for prediction. LSTM and GRU were used to build the model on a training set of 1307 motion sequences and the models were tested on an unseen test data of 289 motion sequences. DTW was used for evaluation. For our architectures three different activation functions (sigmoid/linear/tanh) were used. At first linear activation function was utilized whose value is not closed between any range. I chose linear function, because our task was to predict weight of a container at various time-point in a pouring sequence. So, using linear function would grasp differences between different materials used for pouring. Then experimentation was done with sigmoid and tanh activation function. Sigmoid value lies between 0 and 1. Sigmoid function is differentiable, so slope can be found at any two points in the curve. For this purpose, the test label's range were modified in between 0 and 1. Tanh is another function that was used for analysis, whose value lies in the range of -1 to 1, it can be called as scaled sigmoid. Tanh is better than sigmoid because it can handle negative points efficiently and its curve is also s-shaped so slope between two points in the curve can be obtained. From our prediction plots and dwt plots, we found that the predicted values are almost same as the actual one. Though the intricate changes in weight are not getting generated properly but our approach showed great promise for the learning activity. It is found that, using GRU with tanh function for prediction grasped best behavior in the unseen test data.

Future work involves, experimenting with training for more epochs and change the model architecture to generate a better trajectory.